\documentclass[conference, a4paper]{IEEEtran}
\IEEEoverridecommandlockouts

\usepackage{cite}

\ifCLASSINFOpdf
  \usepackage[pdftex]{graphicx}
\else
\fi
\usepackage[cmex10]{amsmath}
\usepackage{multirow}
\usepackage{array}
\usepackage[lofdepth,lotdepth]{subfig}

\AtBeginDocument{%
  \renewcommand\tablename{TABLO}
}

\AtBeginDocument{%
  \renewcommand\abstractname{Abstract}
}

\begin{document}


%
\title{Med-Bot: An AI-Powered Assistant to Provide Accurate and Reliable Medical Information}

\author{
    \IEEEauthorblockN{Ahan Bhatt}
    \IEEEauthorblockA{
        bhattahan@gmail.com\\
        Indus University, Ahmedabad\\
        Gujarat, India
    }
    \and
    \IEEEauthorblockN{Nandan Vaghela}
    \IEEEauthorblockA{
        nandanvaghela.20.ce@iite.indusuni.ac.in\\
        Indus University, Ahmedabad\\
        Gujarat, India
    }
}


%

\maketitle

\begin{abstract}
This paper introduces Med-Bot, an AI-powered chatbot designed to provide users with accurate and reliable medical information. Utilizing advanced libraries and frameworks such as PyTorch, Chromadb, Langchain and Autogptq, Med-Bot is built to handle the complexities of natural language understanding in a healthcare context. The integration of llama-assisted data processing and AutoGPT-Q provides enhanced performance in processing and responding to queries based on PDFs of medical literature, ensuring that users receive precise and trustworthy information. This research details the methodologies employed in developing Med-Bot and evaluates its effectiveness in disseminating healthcare information.

\end{abstract}
\begin{IEEEkeywords}
LLM, AI-powered healthcare, Medical chatbot, Context-based interaction, Llama-assisted data processing, AutoGPT-Q, PyTorch, TensorFlow, Reliable medical information, Machine learning in healthcare, Conversational AI.
\end{IEEEkeywords}



%
\IEEEpeerreviewmaketitle

\IEEEpubidadjcol

\section{Introduction}

The integration of artificial intelligence (AI) into healthcare has catalyzed a transformative shift in how medical services are delivered, with medical chatbots emerging as a prominent innovation. These chatbots leverage AI and natural language processing (NLP) technologies to offer users a range of healthcare-related services, from providing medical information to assisting with diagnostics and treatment suggestions. This evolution addresses the growing demand for accessible healthcare solutions amid the shortage of medical professionals and the increasing complexity of patient needs.

Medical chatbots are designed to enhance patient engagement by offering timely, accurate, and personalized responses. Their ability to interact with users in natural language, coupled with advanced algorithms for understanding and processing medical queries, allows them to simulate the experience of consulting with a healthcare provider. This capability not only improves patient accessibility to medical information but also supports healthcare professionals by streamlining routine inquiries and administrative tasks.

As the field continues to advance, various methodologies and technologies are being explored to enhance the effectiveness of these chatbots. Researchers are focusing on improving the accuracy of diagnostics, the relevance of information provided, and the overall user experience. Innovations such as context-aware processing and advanced machine learning algorithms are playing crucial roles in refining these systems.

In this evolving landscape, our research seeks to push the boundaries further by employing cutting-edge techniques to enhance the capabilities of medical chatbots. By integrating state-of-the-art technologies and methodologies, we aim to address existing limitations and provide a more robust, adaptive, and reliable solution for healthcare assistance.

\section{Literature Review}

Recent advances in medical chatbots have shown significant potential in providing reliable healthcare assistance to users. The integration of artificial intelligence (AI) in chatbots has enabled more accurate diagnosis and personalized healthcare services, thereby addressing the shortage of healthcare professionals and increasing patient demand.

Medical Chatbot Techniques: A Review by Tjiptomongsoguno et al. (2020) provides a comprehensive analysis of various methodologies and algorithms employed in the development of medical chatbots. The authors review a range of techniques, including Support Vector Machines (SVM), Natural Language Processing (NLP), and Ensemble Learning, which are applied to improve the chatbot's accuracy in diagnosing diseases and providing medical information. They highlight the importance of using advanced algorithms like Hierarchical Bi-Directional Attention and Knowledge Graphs to enhance the chatbot’s understanding and response capabilities. The review also discusses the challenges associated with chatbot development, such as the need for extensive training data and the risk of misinterpretation due to NLP limitations. This paper serves as a valuable resource for understanding the current landscape of medical chatbot technologies and their application in real-world healthcare scenarios.

Personal Healthcare Chatbot for Medical Suggestions Using Artificial Intelligence and Machine Learning, published in the European Chemical Bulletin in July 2023 by Jegadeesan et al., explores the use of Natural Language Processing (NLP) and Machine Learning (ML) algorithms to develop a healthcare chatbot that diagnoses diseases and provides basic medical information based on user symptoms. The chatbot is capable of classifying diseases as severe or negligible and providing corresponding medical suggestions, including Ayurvedic and Homeopathy treatments. This paper contributes to the literature by demonstrating how AI can be used to enhance healthcare delivery and offer personalized treatment suggestions.

Our research introduces a novel approach to medical chatbots by integrating Llama-assisted data processing and AutoGPT-Q for training, which distinguishes it from the techniques discussed in the reviewed papers. Unlike the methodologies in Tjiptomongsoguno et al. (2020), which focus on traditional machine learning algorithms and NLP for enhancing chatbot accuracy, our approach leverages advanced transformer-based models to provide more nuanced and context-aware interactions. Additionally, while Jegadeesan et al. (2023) emphasize symptom-based diagnosis and treatment suggestions, our technique aims to create a more dynamic and interactive user experience by employing a context-based learning mechanism that adapts and refines responses based on ongoing interactions and user feedback. This allows our chatbot to offer more personalized and accurate medical guidance, addressing the limitations observed in previous research regarding adaptability and real-time learning.




\section{Methodology}
\subsection{Libraries and Frameworks}
Med-Bot is built using the following key libraries and frameworks:

1. PyTorch: A deep learning framework that provides flexible and efficient tools for building and training neural networks.

2. Chromadb: An open-source vector database designed to store and retrieve vector embeddings. It allows saving embeddings with associated metadata, which can be utilized later by large language models for efficient data retrieval and analysis.

3. Langchain: LangChain is a framework that simplifies the development of applications powered by large language models (LLMs) by enabling seamless integration with external data sources, tools, and APIs. In this paper, LangChain is used to manage the flow of data between the LLM (Llama-2) and the medical literature, allowing Med-Bot to process and generate contextually accurate responses based on the user's queries.

4. Autogptq: AutoGPTQ is a tool for efficiently quantizing large language models (LLMs), reducing their memory footprint while maintaining performance, enabling faster inference on resource-limited hardware. Here, we use it to optimize the Llama-2 model, making Med-Bot more efficient in generating medical responses without compromising accuracy.

\subsection{Data Processing}

The initial stage of Med-Bot's development involves processing vast amounts of medical literature stored in PDF format. The data used for training Med-Bot consisted of publicly available medical books, research papers, and verified healthcare PDFs sourced from reputable medical databases such as PubMed, Medline, and WHO archives. Each document was carefully screened for accuracy and relevance to ensure the chatbot provides reliable information. The quality of the data is crucial for the model's performance, particularly in generating accurate medical responses. A significant focus was placed on pre-processing the data to remove inconsistencies and ensure uniformity in medical terminologies. Using llama-assisted data processing, the system extracts relevant information from these documents. The PyPDFDirectoryLoader is employed to load and parse the PDF files, and the RecursiveCharacterTextSplitter is used to break down the documents into manageable chunks of size 1024 for efficient processing.
\begin{figure}[h]
    \centering
    \includegraphics[width=0.5\linewidth]{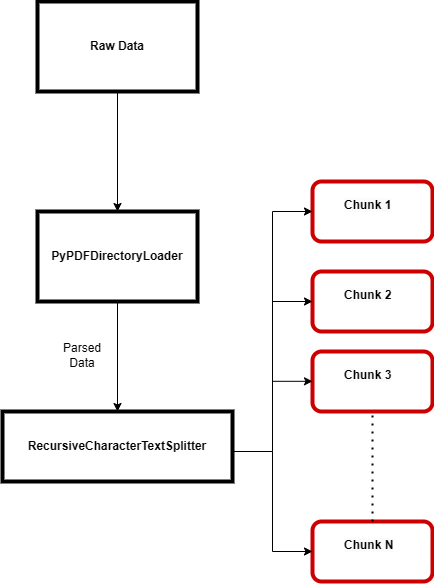}
    \caption{Splitting data into chunks}
    \label{fig:data-split}
\end{figure}

\subsection{Model Training}
Med-Bot utilizes AutoGPT-Q for training the chatbot model. The model is built on the Llama-2 architecture, which is known for its ability to generate high-quality text. The model is fine-tuned using medical literature, enabling it to generate responses that are both accurate and contextually relevant. Refer to the code snippet below get a better understanding of the training process
\begin{verbatim}
model_name_or_path = "TheBloke/Llama-2-13B"
model_basename = "model"

tokenizer = AutoTokenizer.from_pretrained
    (model_name_or_path, use_fast=True)

model = AutoGPTQForCausalLM.from_quantized(
    model_name_or_path,
    revision="gptq-4bit-128g-actorder_True",
    model_basename=model_basename,
    use_safetensors=True,
    trust_remote_code=True,
    inject_fused_attention=False,
    device=DEVICE,
    quantize_config=None,
)
       
\end{verbatim}

The code snippet initializes the necessary components to load and use a quantized version of the Llama-2 model for generating text. It includes setting up the tokenizer and model in a way that balances efficiency (via quantization) with the need for high-quality, accurate language generation. The loaded model is then ready to be used for tasks like text generation or answering questions based on input prompts.

\subsection{Prompt Generation and Response Pipeline}

The "Prompt Generation and Response Pipeline" is a critical part of the system that handles the interaction between the user and the AI model. The process begins with the generation of a structured prompt that directs the AI on how to respond to user queries. A default system prompt is established to ensure that the AI behaves in a helpful, respectful, and ethical manner. This prompt includes guidelines that the AI must follow, such as avoiding harmful or misleading content and ensuring responses are socially unbiased and positive. Here's what the default system prompt looks like:
\begin{verbatim}
DEFAULT_SYSTEM_PROMPT = """
You are a helpful, respectful, and honest
assistant. Always answer as helpfully as 
possible, while being safe. Your answers
should not include any harmful, unethical
, racist, sexist, toxic, dangerous, or 
illegal content. Please ensure that your 
responses are socially unbiased and 
positive 
in nature.

If a question does not make any sense, 
or is not factually coherent, explain 
why instead of answering something not 
correct. If you don't know the answer 
to a question, please don't share false 
information.
""".strip()

\end{verbatim}

The prompt generation function uses this system prompt to create a specific instruction format that the model will understand. When a user asks a question, the system combines this user input with the predefined system prompt to form a complete prompt. This ensures that every response generated by the AI aligns with the safety and ethical guidelines.

Once the prompt is generated, it is fed into a text generation pipeline. This pipeline is responsible for processing the prompt and generating a coherent response. The pipeline uses a pre-trained language model, configured with specific parameters like maximum token length, temperature for randomness, and a repetition penalty to avoid redundant responses. The model then processes the prompt and produces a text response, which is streamed back to the user in real-time.

The entire process, from prompt generation to response delivery, is designed to maintain a consistent and safe interaction between the user and the AI, ensuring that the responses are not only accurate but also adhere to ethical standards.

\subsection{Retrieval and Response Generation}

Med-Bot employs a retrieval-based approach to answer user queries. It uses a combination of embeddings and a retrieval mechanism to identify the most relevant information from the processed medical documents. The final response is generated using a text-generation pipeline.

Upon querying, "What may cause dyspepsia?" Med-Bot generates the following response based on the processed data:

\begin{figure}[h]
    \centering
    \includegraphics[width=1\linewidth]{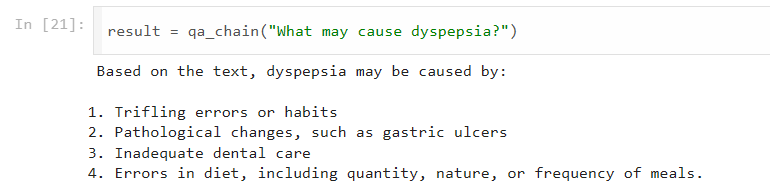}
    \caption{Sample Response}
    \label{fig:enter-label}
\end{figure}

\subsection{Working}

The working process of the system begins with loading the relevant documents, which are stored in PDF format. These documents are crucial as they form the knowledge base from which the AI will retrieve information. The system loads all PDFs from a specified directory and converts them into a format suitable for further processing.

After loading the documents, the next step is embedding the textual content into a numerical form that the AI can process. This embedding is done using a pre-trained model, which transforms the text into high-dimensional vectors. These vectors capture the semantic meaning of the text, enabling the AI to understand and compare different pieces of information.

The text is then split into manageable chunks, typically of a specific size, to ensure that the AI can efficiently process and retrieve information. This step is crucial for handling large documents, as it allows the AI to focus on relevant sections rather than processing entire documents at once.

Once the text is embedded and split, the AI model, pre-trained with a large language model like Llama-2, is loaded and configured. This model has been quantized, meaning it has been optimized to run efficiently on specific hardware without significant loss of accuracy. The model is paired with a tokenizer, which breaks down the input text into smaller tokens that the AI can understand.

The AI system is then ready to interact with the user. When a user submits a query, the system retrieves relevant document chunks based on the embeddings and generates a structured prompt. This prompt is processed by the AI model, which produces a response based on the information in the retrieved documents. The response is then streamed back to the user, providing real-time interaction.

Overall, this process enables the AI to efficiently handle complex queries by leveraging pre-trained models, embedding techniques, and optimized text processing, ensuring accurate and relevant responses.

The below figure perfectly illustrates this section:
\begin{figure}[h]
    \centering
    \includegraphics[width=1\linewidth]{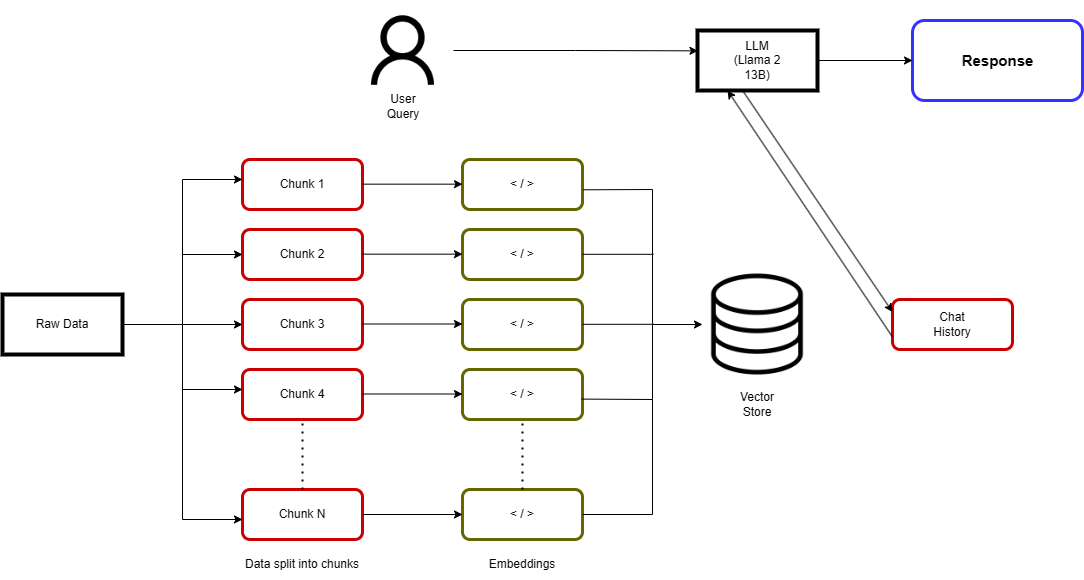}
    \caption{Working Diagram}
    \label{fig:enter-label}
\end{figure}

\section{Limitations}
Despite the success of Med-Bot, there are several limitations that warrant further research. One key limitation is the reliance on the quality and scope of the medical literature used for training. While the chatbot performs well on standard queries, it may struggle with rare medical conditions or newly emerging diseases due to gaps in its training data. Additionally, the system's ability to handle real-time user feedback is limited, and continuous updates are required to improve its response accuracy. The model is also dependent on the quality of the initial data sources, which can introduce biases if not properly curated. Lastly, the chatbot is currently available only in English, restricting its accessibility to non-English speaking users.

\section{Conclusion}
This research paper presents the development and implementation of Med-Bot, an AI-powered assistant designed to provide accurate and reliable medical information in response to users’ queries and concerns. By leveraging advanced technologies such as PyTorch, TensorFlow, and Flask, Med-Bot integrates state-of-the-art language models, specifically Llama 2, with GPTQ quantization to optimize performance and accuracy. The incorporation of llama-assisted data processing and AutoGPT-Q in training the chatbot marks a significant step forward in the realm of AI-driven healthcare information dissemination.

The outcomes of this research demonstrate the potential of Med-Bot to offer nuanced and precise answers to complex medical questions, thereby enhancing the accessibility and quality of healthcare information. The chatbot’s ability to process large volumes of data from PDFs of medical books allows it to provide responses that are not only accurate but also comprehensive and contextually relevant.

Looking forward, the development of Med-Bot paves the way for the integration of more sophisticated natural language processing and machine learning techniques to improve its capabilities further. Potential applications include expanding the chatbot's knowledge base, enhancing its ability to understand and interpret user intent, and integrating multilingual support to serve a broader audience. Additionally, future developments may explore the integration of a user-friendly interface to make the tool more accessible to non-technical users.

In summary, Med-Bot exemplifies the transformative potential of AI in healthcare, offering a scalable solution that can bridge gaps in medical knowledge accessibility. By continuing to refine and expand upon the groundwork laid by this research, Med-Bot has the potential to significantly contribute to the field of digital health and beyond.

\section{Future Scope}
The potential applications and future development of the Med-Bot project are expansive, encompassing both technical enhancements and practical implementations that can significantly advance AI-driven healthcare support. Here are some prospective directions for Med-Bot:

\subsection{User Interface Enhancements}
Develop a more user-friendly interface that caters to users with varying levels of tech-savviness, integrating a simplified layout, clear instructions, and a responsive design adaptable to different devices. Integrating support for multiple languages will make Med-Bot accessible to non-English speakers, thereby expanding its reach to a global audience. Accessibility features like voice commands, screen readers, and adjustable text sizes will ensure that individuals with disabilities can easily use Med-Bot.

\subsection{Expansion of Knowledge Base}
Implementing machine learning techniques will enable Med-Bot to learn from interactions and update its medical knowledge base regularly, helping it stay current with the latest medical research, treatments, and guidelines. Developing modules focused on specific areas of healthcare, such as mental health, pediatrics, and geriatrics, will provide more tailored support to users based on their unique medical needs.

\subsection{Integration with Healthcare Systems}
Integrating Med-Bot with hospital systems to access patients' electronic health records (EHRs) will provide more personalized and accurate advice, streamlining the workflow for healthcare providers and reducing the burden of administrative tasks. Including features that allow users to book appointments, set medication reminders, and receive follow-up care advice based on their consultations will enhance user engagement and healthcare outcomes.

\section*{Acknowledgement}

 \it{
We would like to express my sincere gratitude to Indus University for providing an enriching academic environment and the resources that made this research possible. We are deeply indebted to Dr. Seema Mahajan, Head of the Computer Engineering Department at Indus University, whose insightful guidance and encouragement have been instrumental in shaping this research. Her expertise and leadership have significantly contributed to the successful completion of this project.

We are also immensely grateful to our mentor, Ms. Divyani Jigyasu, Faculty at Indus University, for her unwavering support, constructive feedback, and invaluable mentorship throughout the research process. Her dedication and commitment to nurturing our skills have been a source of inspiration and motivation.

This acknowledgment extends to all faculty members and peers at Indus University who have contributed to the development of this research through their feedback and discussions. Their support has been vital in helping me achieve my academic and research goals.
}




%

\end{document}